\documentclass{article}

\usepackage{microtype}
\usepackage{graphicx}
\usepackage{subfigure}
\usepackage{booktabs} 
\usepackage{url}
\usepackage{algorithmic}
\usepackage{algorithm}
\usepackage{color}
\usepackage{rotating}
\usepackage{hhline}
\usepackage{multirow}
\usepackage{amsfonts, amstext, amsmath, amssymb, amsthm}
\usepackage{bbm}
\usepackage{bm}
\usepackage[inline]{enumitem}
\usepackage[title]{appendix}
\usepackage{colortbl}
\usepackage{picture}
\usepackage{tikz}
\usetikzlibrary{shapes}
\usepackage{multirow}
\usepackage{tabu}
\usepackage{adjustbox}
\usepackage{nicefrac}

\usepackage[nohyperref,accepted]{arxiv_icml2019} 

\icmltitlerunning{Geometric Scattering for Graph Data Analysis}

\newcommand{\zerob}{\bm{0}}
\newcommand{\oneb}{\bm{1}}
\newcommand{\phib}{\bm{\phi}}
\newcommand{\varphib}{\bm{\varphi}}
\newcommand{\xb}{\mathbf{x}}
\newcommand{\vb}{\mathbf{v}}
\newcommand{\Psib}{\bm{\Psi}}

\newcommand{\db}{\mathbf{d}}
\newcommand{\mub}{\bm{\mu}}
\newcommand{\yb}{\mathbf{y}}

\newcommand{\R}{\mathbb{R}}

\newcommand{\Ab}{\mathbf{A}}
\newcommand{\Db}{\mathbf{D}}

\newcommand{\Ib}{\mathbf{I}}
\newcommand{\Lb}{\mathbf{L}}

\newcommand{\Nb}{\mathbf{N}}
\newcommand{\Pb}{\mathbf{P}}

\begin{document}

\twocolumn[
\icmltitle{Geometric Scattering for Graph Data Analysis}



\icmlsetsymbol{equal}{*}

\begin{icmlauthorlist}
\icmlauthor{Feng Gao}{CMSE,PSMS}
\icmlauthor{Guy Wolf}{udem}
\icmlauthor{Matthew Hirn}{CMSE,math}
\end{icmlauthorlist}

\icmlaffiliation{CMSE}{Department of Computational Math, Science and Engineering,}
\icmlaffiliation{PSMS}{Department of Plant, Soil \& Microbial Sciences,}
\icmlaffiliation{math}{Department of Mathematics, Michigan State University, East Lansing, MI, USA}
\icmlaffiliation{udem}{Department of Mathematics and Statistics, Universit\'{e} de Montr\'{e}al, Montreal, QC, Canada}

\icmlcorrespondingauthor{Guy Wolf}{guy.wolf@umontreal.ca}
\icmlcorrespondingauthor{Matthew Hirn}{mhirn@msu.edu}

\icmlkeywords{Geometric deep learning, scattering transform, graph signal processing}

\vskip 0.3in
]



\printAffiliationsAndNotice{}  

\begin{abstract}
We explore the generalization of scattering transforms from traditional (e.g., image or audio) signals to graph data, analogous to the generalization of ConvNets in geometric deep learning, and the utility of extracted graph features in graph data analysis. In particular, we focus on the capacity of these features to retain informative variability and relations in the data (e.g., between individual graphs, or in aggregate), while relating our construction to previous theoretical results that establish the stability of similar transforms to families of graph deformations. We demonstrate the application the our geometric scattering features in graph classification of social network data, and in data exploration of biochemistry data.
\end{abstract}

\vspace{-20pt}

\section{Introduction}

Over the past decade, numerous examples have established that deep neural networks (i.e., cascades of linear operations and simple nonlinearities) typically outperform traditional ``shallow'' models in various modern machine learning applications, especially given the increasing Big Data availability nowadays. Perhaps the most well known example of the advantages of deep networks is in computer vision, where the utilization of 2D convolutions enable network designs that learn cascades of convolutional filters, which have several advantages over fully connected network architectures, both computationally and conceptually. Indeed, in terms of supervised learning, convolutional neural networks (ConvNets) hold the current state of the art in image classification, and have become the standard machine learning approach towards processing big structured-signal data, including audio and video processing. See, e.g., \citet[Chapter 9]{Goodfellow-et-al-2016} for a detailed discussion. 

Beyond their performances when applied to specific tasks, pretrained ConvNet layers have been explored as image feature extractors by freezing the first few pretrained convolutional layers and then retraining only the last few layers for specific datasets or applications~\citep[e.g.,][]{Yosinski:deepLearnTransfer2014,oquab2014learning}. Such transfer learning approaches provide evidence that suitably constructed deep filter banks should be able to extract task-agnostic semantic information from structured data, and in some sense mimic the operation of human visual and auditory cortices, thus supporting the neural terminology in deep learning. An alternative approach towards such universal feature extraction was presented in \citet{mallat:scattering2012}, where a deep filter bank, known as the scattering transform, is \emph{designed}, rather than trained, based on predetermined families of distruptive patterns that should be eliminated to extract informative representations. The scattering transform is constructed as a cascade of linear wavelet transforms and nonlinear complex modulus operations that provides features with guaranteed invariance to a predetermined Lie group of operations such as rotations, translations, or scaling. Further, it also provides Lipschitz stability to small diffeomorphisms of the inputted signal.

Following recent interest in geometric deep learning approaches for processing graph-structured data (see, for example, \citet{Bronstein:geoDeepLearn2017} and references therein), several attempts have been made to generalize the scattering transform to graphs \citep{zou:graphCNNScat2018, gama:diffScatGraphs2018} and manifolds \citep{perlmutter:geoScatManifolds2018}, which we will generally term ``geometric scattering''. These works mostly focus on following the footsteps of \citet{mallat:scattering2012} in establishing the stability of their respective constructions to deformations of input signals or graphs. Their results essentially characterize the type of disruptive information eliminated by geometric scattering, by providing upper bounds for distances between scattering features, phrased in terms of a deformation size. Here, we further explore the notion of geometric scattering features by considering the complimentary question of how much information is retained by them, since stability alone does not ensure useful features in practice (e.g., a constant all-zero map would be stable to any deformation, but would clearly be useless). In other words, we examine whether a geometric scattering construction, defined and discussed in Sec. \ref{sec: graph scat}, can be used as an effective task-independent feature extractor from graphs, and whether the resulting representations provided by them are sufficiently rich to enable intelligible data analysis by applying traditional (Euclidean) methods. 

We note that for Euclidean scattering, while stability is established with rigorous theoretical results, the capacity of scattering features to form an effective data representation in practice has mostly been established via extensive empirical examination. Indeed, scattering features have been shown effective in several audio~\citep[e.g.,][]{bruna:audioTextureSynth2013, anden:deepScatSpectrum2014, lostanlen:scatPitchSpiral2015, arXiv:1807.08869} and image~\citep[e.g.,][]{bruna:invariantScatConvNet2013, mallat:rigidMotionScat2014, oyallon:scatObjectClass2014, angles2018generative} processing applications, and their advantages over learned features are especially relevant in applications with relatively low data availability, such as quantum chemistry and materials science~\citep[e.g.,][]{hirn:waveletScatQuantum2016, eickenberg:3DSolidHarmonicScat2017, eickenberg:scatMoleculesJCP2018, brumwell:steerableScatLiSi2018}. 

Similarly, our examination of geometric scattering capacity focuses on empirical results on several data analysis tasks, and on two commonly used graph data types. Our results in Sec.~\ref{subsec: classification} show that on social network data, geometric scattering features enable classic RBF-kernel SVM to match, if not outperform, leading graph kernel methods as well as most geometric deep learning ones. These experiments are augmented by additional results in Sec.~\ref{subsec: low train} that show the geometric scattering SVM classification rate degrades only slightly when trained on far fewer graphs than is traditionally used in graph classification tasks. On biochemistry data, where graphs represent molecular structures of compounds (e.g., Enzymes or proteins), we show in Sec.~\ref{subsec: dim red} that scattering features enable significant dimensionality reduction. Finally, to establish their descriptive qualities, in Sec.~\ref{subsec: exploration} we use geometric scattering features extracted from enzyme data~\citep{borgwardt:ENZYMES} to infer emergent patterns of enzyme commission (EC) exchange preferences in enzyme evolution, validated with established knowledge from~\citet{cuesta:EC-evolution}. Taken together, these results illustrate the power of the geometric scattering approach as both a relevant mathematical model for geometric deep learning, and as a suitable tool for modern graph data analysis.

\section{Graph Random Walks and Graph Wavelets} \label{sec: graph wavelets}

The Euclidean scattering transform is constructed using wavelets defined on $\R^d$. In order to extend this construction to graphs, we define graph wavelets as the difference between lazy random walks that have propagated at different time scales, which mimics classical wavelet constructions found in \citet{meyer:waveletsOperators1993} and more recent constructions found in \citet{coifman:diffWavelets2006}. The underpinnings for this construction arise out of graph signal processing, and in particular the properties of the graph Laplacian.

Let $G = (V, E, W)$ be a weighted graph, consisting of $n$ vertices $V = \{v_1, \ldots, v_n\}$, edges $E \subseteq \{ (v_{\ell}, v_m)  : 1 \leq \ell,m \leq n \}$, and weights $W = \{ w (v_{\ell}, v_m) > 0 : (v_{\ell}, v_m) \in E \}$. Note that unweighted graphs are considered as a special case, by setting $w (v_{\ell}, v_m) = 1$ for each $(v_{\ell}, v_m) \in E$. Define the $n \times n$ (weighted) adjacency matrix $\Ab_G = \Ab$ of $G$ by $\Ab (v_{\ell}, v_m) = w(v_{\ell}, v_m)$ if $(v_{\ell}, v_m) \in E$ and zero otherwise, where we use the notation $\Ab (v_{\ell}, v_m)$ to denote the $(\ell,m)$ entry of the matrix $\Ab$ so as to emphasize the correspondence with the vertices in the graph and to reserve sub-indices for enumerating objects. Define the (weighted) degree of vertex $v_{\ell}$ as $\deg (v_{\ell}) = \sum_m \Ab (v_{\ell}, v_m)$ and the corresponding diagonal $n \times n$ degree matrix $\Db$ given by $\Db (v_{\ell}, v_{\ell}) = \deg (v_{\ell})$, $\Db (v_{\ell}, v_m) = 0$, $\ell \neq m$. Finally, the $n \times n$ graph Laplacian matrix $\Lb_G = \Lb$ on $G$ is defined as $\Lb = \Db - \Ab$, and its normalized version is $\Nb = \Db^{-\nicefrac{1}{2}} \Lb \Db^{-\nicefrac{1}{2}} = \Ib - \Db^{-\nicefrac{1}{2}} \Ab \Db^{-\nicefrac{1}{2}}$. We focus on the latter due to its close relationship with graph random walks.

The normalized graph Laplacian is a symmetric, real valued positive semi-definite matrix, and thus has $n$ non-negative eigenvalues. Furthermore, if we set $\zerob = (0, \ldots, 0)^T$ to to be the $n \times 1$ vector of all zeroes, and $\db (v_{\ell}) = \deg (v_{\ell})$ to be the $n \times 1$ degree vector, 
then one has $\Nb \db^{\nicefrac{1}{2}} = \zerob$ (where the square root is understood to be taken entrywise). Therefore $0$ is an eigenvalue of $\Nb$ and we write the $n$ eigenvalues of $\Nb$ as $0 = \lambda_0 \leq \lambda_1 \leq \cdots \leq \lambda_{n-1} \leq 2$ with corresponding $n \times 1$ orthonormal eigenvectors $\varphib_0, \varphib_1, \ldots, \varphib_{n-1}$. If the graph $G$ is connected, then $\lambda_1 > 0$. In order to simplify the following discussion we assume that this is the case, although the discussion below can be amended to include disconnected graphs as well.

One can show $\varphib_0 = \db^{\nicefrac{1}{2}} / \| \db^{\nicefrac{1}{2}} \|$, meaning $\varphib_0$ is non-negative. Since every other eigenvector is orthogonal to $\varphib_0$ (and thus must take positive and negative values), it is natural to view the eigenvectors $\varphib_k$ as the Fourier modes of the graph $G$, with a frequency magnitude proportional to $\lambda_k$. The fact that $\varphib_0$ is in general non-constant, as opposed to the zero frequency mode on the torus or real line, reflects the non-uniform distribution of vertices in non-regular graphs. Let $\xb : V \rightarrow \R$ be a signal defined on the vertices of the graph $G$, which we will consider as an $n \times 1$ vector with entries $\xb (v_{\ell})$. It follows that the Fourier transform of $\xb$ can be defined as $\widehat{\xb} (k) = \xb \cdot \varphib_k$, where $\xb \cdot \mathbf{y}$ is the standard dot product. This analogy is one of the foundations of graph signal processing and indeed we could use this correspondence to define wavelet operators on the graph $G$, as in \citet{hammond:graphWavelets2011}. Rather than follow this path, though, we instead take a related path similar to~\citet{coifman:diffWavelets2006} and~\citet{gama:diffScatGraphs2018} by defining the graph wavelet operators in terms of random walks defined on $G$, which will avoid diagonalizing $\Nb$ and will allow us to control the ``spatial'' graph support of the filters directly.

Define the $n \times n$ lazy random walk matrix as $\Pb = \tfrac{1}{2} \left( \Ib + \Ab \Db^{-1} \right)$. Note that the column sums of $\Pb$ are all one. It follows that $\Pb$ acts as a Markov operator, mapping probability distributions to probability distribution. We refer to $\Pb$ as a lazy random walk matrix since $\Pb^t$ governs the probability distribution of a lazy random walk after $t$ steps. A single realization of a random walk is a walk (in the graph theoretic sense) $v_{\ell_0}, v_{\ell_1}, v_{\ell_2}, \ldots$ in which the steps are chosen randomly; lazy random walks allow for $v_{\ell_i} = v_{\ell_{i+1}}$. More precisely, suppose that $\mub_0 (v_\ell) \geq 0$ for each vertex $v_{\ell}$ and $\| \mub_0 \|_1 = 1$, so that $\mub_0$ is a probability distribution on $G$. We take $\mub_0 (v_{\ell})$ as the probability of a random walk starting at vertex $v_{\ell_0} = v_{\ell}$. One can verify that $\mub_1 = \Pb \mub_0$ is also a probability distribution; each entry $\mub_1 (v_{\ell})$ gives the probability of the random walk being located at $v_{\ell_1} = v_{\ell}$ after one step. The probability distribution for the location of the random walk after $t$ steps is $\mub_t = \Pb^t \mub_0$.

The operator $\Pb$ can be considered a low pass operator, meaning that $\Pb \xb$ replaces $\xb (v_{\ell})$ with localized averages of $\xb (v_{\ell})$ for any $\xb$. Indeed, expanding out $\Pb \xb (v_{\ell})$ one observes that $\Pb \xb (v_{\ell})$ is the weighted average of $\xb (v_{\ell})$ and the values $\xb (v_m)$ for the neighbors $v_m$ of $v_{\ell}$. Similarly, the value $\Pb^t \xb (v_{\ell})$ is the weighted average of $\xb (v_{\ell})$ with all values $\xb (v_m)$ such that $v_m$ is within $t$ steps of $v_{\ell}$.

Low pass operators defined on Euclidean space retain the low frequencies of a function while suppressing the high frequencies. The random walk matrix $\Pb$ behaves similarly. Indeed, $\Pb$ is diagonalizable with $n$ eigenvectors $\phib_k = \Db^{\nicefrac{1}{2}} \varphib_k$ and eigenvalues $\omega_k = 1 - \nicefrac{\lambda_k}{2}$. Let $\mathbf{y}_{\xb} = \Db^{-\nicefrac{1}{2}} \xb$ be a density normalized version of $\xb$ and set $\xb_t = \Pb^t \xb$; then one can show
\begin{equation} \label{eqn: P is a low pass}
    \yb_{\xb_t} = \widehat{\yb_{\xb}}(0) \varphib_0 + \sum_{k=1}^{n-1} \omega_k^t \widehat{\yb_{\xb}}(k) \varphib_k \, .
\end{equation}
Thus, since $0 \leq \omega_k < 1$ for $k \geq 1$, the operator $\Pb^t$ preserves the zero frequency of $\xb$ while suppressing the high frequencies, up to a density normalization. 

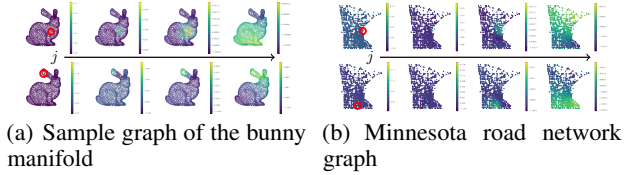
\begin{figure}
    \centering
    \subfigure[Sample graph of the bunny manifold]{\adjustbox{width=0.48\linewidth}{\begin{tabular}{c}
    \includegraphics[width=2.475in,keepaspectratio]{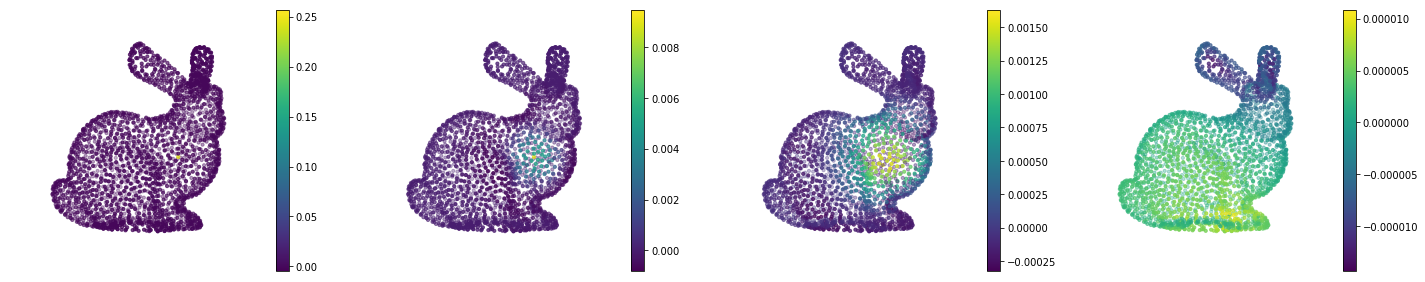}\\
    \includegraphics[width=2.475in,keepaspectratio]{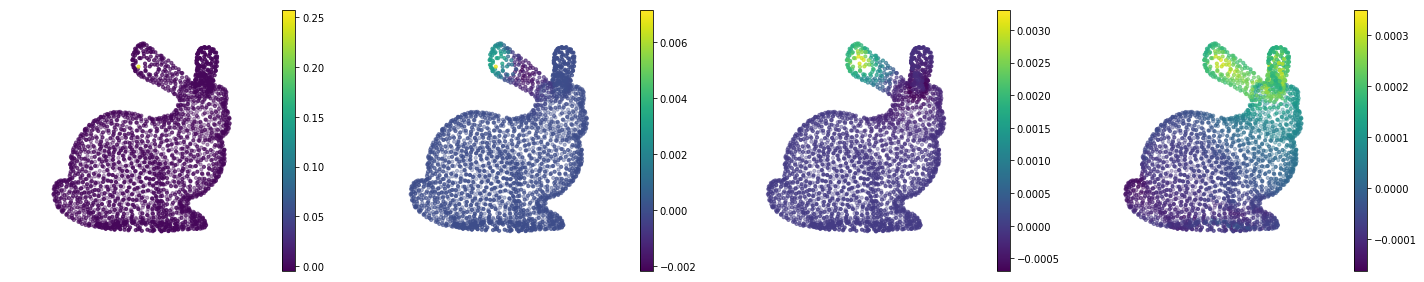}
    \end{tabular} \begin{tikzpicture}[overlay]
    \draw[very thick,red] (-5.80,0.75) circle(0.07);
    \draw[very thick,red] (-5.97,-0.20) circle(0.08);
    \draw[thick,->] (-5.5,0.17) node[left]{$j$} -- (-0.7,0.17);
    \end{tikzpicture}}}
    \,
    \subfigure[Minnesota road network graph]{\adjustbox{width=0.48\linewidth}{\begin{tabular}{c}
    \includegraphics[width=2.475in,keepaspectratio]{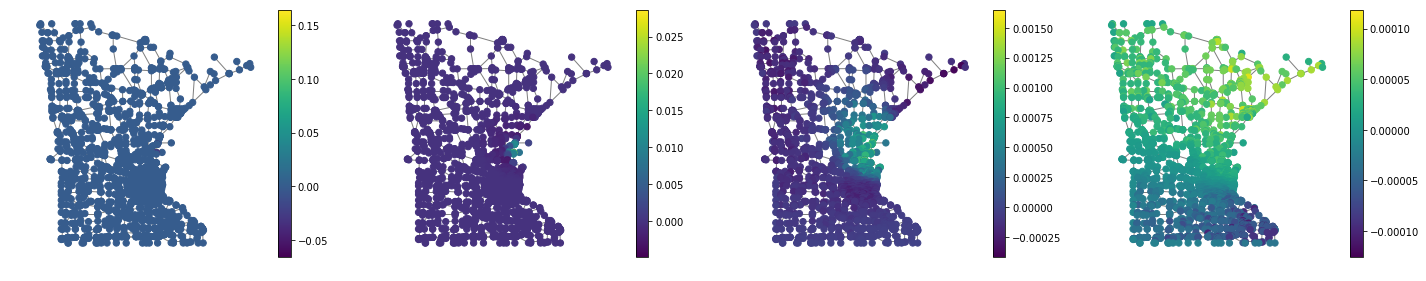}\\
    \includegraphics[width=2.475in,keepaspectratio]{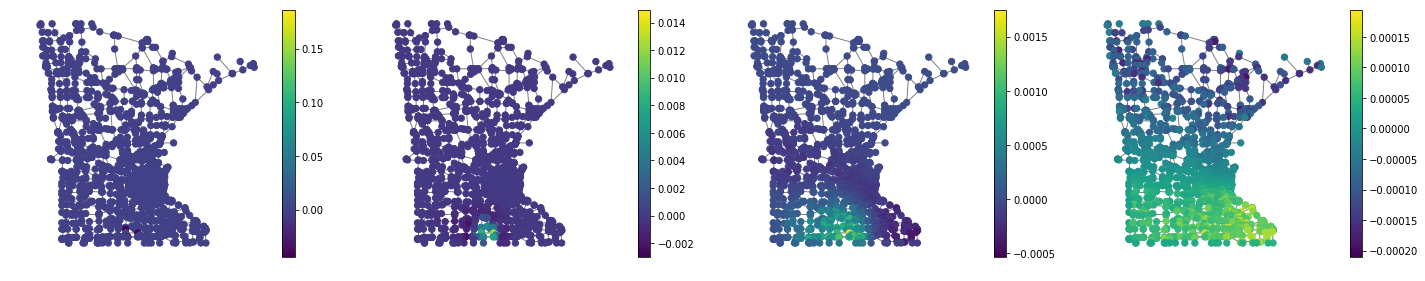}
    \end{tabular} \begin{tikzpicture}[overlay]
    \draw[very thick,red] (-5.89,0.77) circle(0.07);
    \draw[very thick,red] (-6.01,-0.94) circle(0.07);
    \draw[thick,->] (-5.5,0.17) node[left]{$j$} -- (-0.7,0.17);
    \end{tikzpicture}}}
    \caption{Wavelets $\Psib_j$ for increasing scale $2^j$ left to right, applied to Diracs centered at two different locations (marked by red circles) in two different graphs. Vertex colors indicate wavelet values (corresponding to colorbars for each plot), ranging from yellow/green indicating positive values to blue indicating negative values. Both graphs are freely available from~\citet{PyGSP}.}
    \label{fig:wavelets}
\end{figure}

High frequency responses of $\xb$ can be recovered in multiple different fashions, but we utilize multiscale wavelet transforms that group the non-zero frequencies of $G$ into approximately dyadic bands. As shown in~\citet[Lemma 2.12]{mallat:scattering2012}, wavelet transforms are provably stable operators in the Euclidean domain, and the proof of~\citet[Theorem 5.1]{zou:graphCNNScat2018} indicates that similar results on graphs may be possible. Furthermore, the multiscale nature of wavelet transforms will allow the resulting geometric scattering transform (Sec.~\ref{sec: graph scat}) to traverse the entire graph $G$ in one layer, which is valuable for obtaining global descriptions of $G$. Following \citet{coifman:diffWavelets2006}, define the $n \times n$ wavelet matrix at the scale $2^j$ as
\begin{equation} \label{eqn: wavelet def}
\Psib_j = \Pb^{2^{j-1}} - \Pb^{2^j} = \Pb^{2^{j-1}} ( \mathbf{I} - \Pb^{2^{j-1}}) \, .
\end{equation}
A similar calculation as the one required for \eqref{eqn: P is a low pass} shows that $\Psib_j \xb$ partially recovers $\widehat{\yb_{\xb}}(k)$ for $k \geq 1$. The value $\Psib_j \xb (v_{\ell})$ aggregates the signal information $\xb (v_m)$ from the vertices $v_m$ that are within $2^j$ steps of $v_{\ell}$, but does not average the information like the operator $\Pb^{2^j}$. Instead, it responds to sharp transitions or oscillations of the signal $\xb$ within the neighborhood of $v_{\ell}$ with radius $2^j$ (in terms of the graph path distance). The smaller the wavelet scale $2^j$, the higher the frequencies $\Psib_j \xb$ recovers in $\xb$. The wavelet coefficients up to the scale $2^J$ are:
\begin{equation} \label{eqn: diffusion wavelet transform}
\Psib^{(J)} \xb (v_\ell) = \left[\Psib_j \xb (v_\ell) : 1 \leq j \leq J \right].
\end{equation}
Figure \ref{fig:wavelets} plots the wavelets on two different graphs. 

\section{Geometric Scattering on Graphs} \label{sec: graph scat}

A geometric wavelet scattering transform follows a similar construction as the (Euclidean) wavelet scattering transform of \citet{mallat:scattering2012}, but leverages a graph wavelet transform. In this paper we utilize the wavelet transform defined in \eqref{eqn: diffusion wavelet transform} of the previous section, but remark that in principle any graph wavelet transform could be used \citep[see, e.g.,][]{zou:graphCNNScat2018}. In Sec.~\ref{sec: graph scat def} we define the graph scattering transform, in Sec.~\ref{sec: capacity} we discuss its relation to other recently proposed graph scattering constructions \citep{gama:diffScatGraphs2018,zou:graphCNNScat2018}, and in Sec.~\ref{sec: comparison} we describe several of its desirable properties as compared to other geometric deep learning algorithms on graphs.

\subsection{Geometric scattering definitions}
\label{sec: graph scat def}

Machine learning algorithms that compare and classify graphs must be invariant to graph isomorphism, i.e., re-indexations of the vertices and corresponding edges. A common way to obtain invariant graph features is via summation operators, which act on a signal $\xb = \xb_G$ that can be defined on any graph $G$, e.g., $\xb (v_{\ell}) = \deg (v_{\ell})$. The geometric scattering transform, which is described in the remainder of this section, follows such an approach.

The simplest summation operator computes the sum of the responses of the signal $\xb$. As described in \citet{verma2008:GCAPS-CNN}, this invariant can be complemented by higher order summary statistics of $\xb$, the collection of which are statistical moments, and which are also referred to as ``capsules'' in that work. For example, the unnormalized $q^{\text{th}}$ moments of $\xb$ yield the following ``zero'' order scattering moments:
\begin{equation} \label{eqn: zero layer}
S \xb (q) = \sum_{\ell=1}^n \xb (v_{\ell})^q, \quad 1 \leq q \leq Q
\end{equation}
We can also replace \eqref{eqn: zero layer} with normalized (i.e., standardized) moments of $\xb$, in which case we store its mean ($q=1$), variance ($q=2$), skew ($q=3$), kurtosis ($q=4$), and so on. In what follows we discuss the unnormalized moments since their presentation is simpler. The invariants $S \xb (q)$ do not capture the full variability of $\xb$ and hence the graph $G$ upon which the signal $\xb$ is defined. We thus complement these moments with summary statistics derived from the wavelet coefficients of $\xb$, which will lead naturally to the graph ConvNet structure of the geometric scattering transform.

Observe, analogously to the Euclidean setting, that in computing $S \xb (1)$, which is the summation of $\xb (v_{\ell})$ over $V$, we have captured the zero frequency of $\yb_{\xb} = \Db^{-\nicefrac{1}{2}} \xb$ since $\sum_{\ell=1}^n \xb (v_{\ell}) = \xb \cdot \oneb = \yb_{\xb} \cdot \db^{\nicefrac{1}{2}} = \| \db^{\nicefrac{1}{2}} \| \widehat{\yb_{\xb}}(0)$. Higher order moments of $\xb$ can incorporate the full range of frequencies in $\xb$, e.g. $S \xb (2) = \sum_{\ell=1}^n \xb (v_{\ell})^2 = \sum_{k=1}^n \widehat{\xb} (k)^2$, but they are mixed into one invariant coefficient. We can separate and recapture the high frequencies of $\xb$ by computing its wavelet coefficients $\Psib^{(J)} \xb$, which were defined in \eqref{eqn: diffusion wavelet transform}. However, $\Psib^{(J)} \xb$ is not invariant to permutations of the vertex indices; in fact, it is equivariant. Before summing the individual wavelet coefficient vectors $\Psib_j \xb$, though, we must first apply a pointwise nonlinearity. Indeed, define $\oneb = (1, \ldots, 1)^T$ to be the $n \times 1$ vector of all ones, and note that $\Pb^T \oneb = \oneb$, meaning that $\oneb$ is a left eigenvector of $\Pb$ with eigenvalue $1$. It follows that $\Psib_j^T \oneb = \zerob$ and thus $\sum_{\ell=1}^n \Psib_j \xb (v_{\ell}) = \Psib_j \xb \cdot \oneb = \oneb^T \Psib_j \xb = 0$.

We thus apply the absolute value nonlinearity, to obtain nonlinear equivariant coefficients $| \Psib^{(J)} \xb| = \{ |\Psib_j \xb| : 1 \leq j \leq J \}$. We use absolute value because it is equivariant to vertex permutations, non-expansive, and when combined with traditional wavelet transforms on Euclidean domains, yields a provably stable scattering transform for $q=1$. Furthermore, initial theoretical results in~\citet{zou:graphCNNScat2018} and~\citet{gama:diffScatGraphs2018} indicate that similar graph based scattering transforms possess certain types of stability properties as well. As in \eqref{eqn: zero layer}, we extract invariant coefficients from $| \Psib_j \xb|$ by computing its moments, which define the first order geometric scattering moments:
\begin{equation} \label{eqn: first layer}
    S \xb (j, q) = \sum_{\ell=1}^n |\Psib_j \xb (v_{\ell})|^q, \; 1 \leq j \leq J, ~ 1 \leq q \leq Q
\end{equation}
These first order scattering moments aggregate complimentary multiscale geometric descriptions of $G$ into a collection of invariant multiscale statistics. These invariants give a finer partition of the frequency responses of $\xb$. For example, whereas $S \xb (2)$ mixed all frequencies of $\xb$, we see that $S \xb (j, 2)$ only mixes the frequencies of $\xb$ captured by $\Psib_j$. 

First order geometric scattering moments can be augmented with second order geometric scattering moments by iterating the graph wavelet and absolute value transforms. These moments are defined as:
\begin{equation} \label{eqn: second layer}
S \xb (j, j', q) = \sum_{\ell=1}^n | \Psib_{j'} |\Psib_j \xb (v_{\ell}) ||^q, \begin{array}{l}1 \leq j < j' \leq J\\ 1 \leq q \leq Q\,,\end{array}
\end{equation}
which consists of reapplying the wavelet transform operator $\Psib^{(J)}$ to each $|\Psib_j \xb|$ and computing the summary statistics of the magnitudes of the resulting coefficients. The intermediate equivariant coefficients $| \Psib_{j'} | \Psib_j \xb ||$ and resulting invariant statistics $S \xb (j, j', q)$ couple two scales $2^j$ and $2^{j'}$ within the graph $G$, creating features that bind patterns of smaller subgraphs within $G$ with patterns of larger subgraphs (e.g., circles of friends of individual people with larger community structures in social network graphs). The transform can be iterated additional times, leading to third order features and beyond, and thus has the general structure of a graph ConvNet.

\newcommand{\tikzmark}[2][~]{\tikz[overlay,remember picture] \node (#2) {#1};}
\newcommand{\DrawBox}{%
  \begin{tikzpicture}[overlay,remember picture]
    \draw[->,out= 50,in=180,distance=3.5cm] (Input.east) to (Zero.west);
    \draw[->] (Zero.east) to (Output.west);
    \draw[->] (Input.east) to (One0.west);
    \draw[->] (One3.east) to (Output.west);
    \draw[->] (Input.east) to (Two0.west);
    \draw[->] (Two6.east) to (Output.west);
    \draw[->] (One0.east) to (One1.west);
    \draw[->] (One1.east) to (One2.west);
    \draw[->] (One2.east) to (One3.west);
    \draw[->] (Two0.east) to node[below,xshift=10pt,yshift=-15pt](wavelet1){$\underbrace{\hspace{110pt}}_{\Psib_j}$} (Two1.west);
    \draw[->] (Two1.east) to (Two2.west);
    \draw[->] (Two2.east) to (Two3.west);
    \draw[->] (Two3.east) to node[below,xshift=10pt,yshift=-15pt](wavelet1){$\underbrace{\hspace{110pt}}_{\Psib_{j^\prime}}$} (Two4.west);
    \draw[->] (Two4.east) to (Two5.west);
    \draw[->] (Two5.east) to node[below,xshift=23.5pt,yshift=-15pt](wavelet1){$\underbrace{\hspace{36pt}}_{1 \leq q \leq Q}$}(Two6.west);

  \end{tikzpicture}
}

\begin{figure*}[!hbt]
\centering
\vspace{-10pt}
\subfigure[Representative zeroth-, first-, and second-order cascades of the geometric scattering transform for an input graph signal $\xb$. 
]{
\adjustbox{width=0.48\linewidth}{\makebox(0.925\textwidth,105){\input{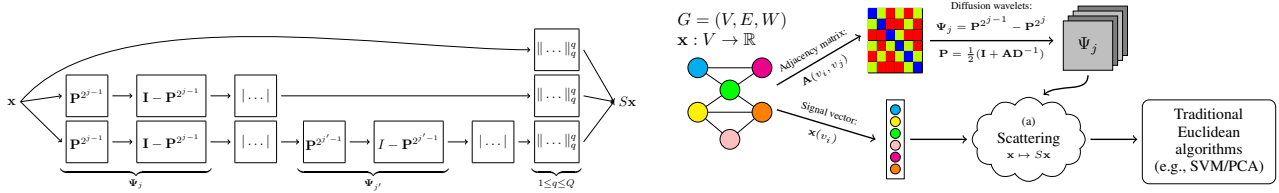}}}
\label{subfig:gscat}}
\hfill 
\subfigure[Architecture for using geometric scattering of graph~$G$ and signal~$\xb$ in graph data analysis, as demonstrated in Sec.~\ref{sec: results}.
]{
\adjustbox{width=0.48\linewidth}{\makebox(0.731\textwidth,110)[t]{\begin{tikzpicture}

\coordinate (t0) at (0,0);

\path (t0)
+(4,1) coordinate (t1)
+(4,-1) coordinate (t2)
+(7.7,1) coordinate (t3)
+(6.7,-1) coordinate (t4)
+(9,-1.9) coordinate (t5);

\path (t0)
+(0,0.45) coordinate (n1) 
+(0,-0.45) coordinate (n2)
+(0.65,0) coordinate (n3)
+(0.65,-1) coordinate (n4)
+(1.3,0.45) coordinate (n5)
+(1.3,-0.45) coordinate (n6);

\path (t2)
+(0,0.60) coordinate (f1) 
+(0,0.36) coordinate (f2)
+(0,0.12) coordinate (f3)
+(0,-0.12) coordinate (f4)
+(0,-0.36) coordinate (f5)
+(0,-0.60) coordinate (f6);


\draw (t0) +(0.7,1.2) node {$\begin{array}{l}
G = (V,E,W)\\
\xb:V \to \mathbb{R}
\end{array}$};

\draw (n1) -- (n3);
\draw (n1) -- (n5);
\draw[black,fill=cyan] (n1) circle(0.2);
\draw (n2) -- (n4);
\draw (n2) -- (n3);
\draw (n2) -- (n6);
\draw[black,fill=yellow] (n2) circle(0.2);
\draw (n3) -- (n5);
\draw (n3) -- (n6);
\draw[black,fill=green] (n3) circle(0.2);
\draw (n4) -- (n6);
\draw[black,fill=pink] (n4) circle(0.2);
\draw[black,fill=magenta] (n5) circle(0.2);
\draw[black,fill=orange] (n6) circle(0.2);


\draw[->,thick] (t1) +(-2.4,-0.9) -- node[midway, above, sloped, align=center]{\tiny{}Adjacency matrix:} node[midway, below, sloped, align=center]{\tiny$\Ab (v_i, v_j)$} +(-0.7,0.1);

\draw (t1) node {\scalebox{0.5}{$\begin{array}{|c|c|c|c|c|c|} 
\hline
\cellcolor{blue} & \cellcolor{lime} & \cellcolor{red} & \cellcolor{lime} & \cellcolor{red} & \cellcolor{lime} \\
\hline
\cellcolor{lime} & \cellcolor{blue} & \cellcolor{red} & \cellcolor{red} & \cellcolor{lime} & \cellcolor{red} \\
\hline
\cellcolor{red} & \cellcolor{red} & \cellcolor{blue} & \cellcolor{lime} & \cellcolor{red} & \cellcolor{red} \\
\hline
\cellcolor{lime} & \cellcolor{red} & \cellcolor{lime} & \cellcolor{blue} & \cellcolor{lime} & \cellcolor{red} \\
\hline
\cellcolor{red} & \cellcolor{lime} & \cellcolor{red} & \cellcolor{lime} & \cellcolor{blue} & \cellcolor{lime} \\
\hline
\cellcolor{lime} & \cellcolor{red} & \cellcolor{red} & \cellcolor{red} & \cellcolor{lime} & \cellcolor{blue} \\
\hline \end{array}$}};

\draw[->,thick] (t2) +(-2.4,0.6) -- node[midway, above, sloped, align=center]{\tiny{Signal vector:}} node[midway, below, sloped, align=center]{\tiny$\xb (v_i)$} +(-0.4,0);
\draw[black] (t2) +(-0.18,-0.8) rectangle +(0.18,0.8);
\draw[black,fill=cyan] (f1) circle(0.1);
\draw[black,fill=yellow] (f2) circle(0.1);
\draw[black,fill=green] (f3) circle(0.1);
\draw[black,fill=pink] (f4) circle(0.1);
\draw[black,fill=magenta] (f5) circle(0.1);
\draw[black,fill=orange] (f6) circle(0.1);


\draw[->,thick] (t3) +(-3,0) -- node[midway, above, sloped, align=center]{\tiny{}Diffusion wavelets:\\\tiny$\Psib_j = \Pb^{2^{j-1}} - \Pb^{2^j}$} node[midway, below, sloped, align=center]{\tiny$\Pb = \frac{1}{2}(\mathbf{I} + \Ab\Db^{-1})$} +(-0.5,0);

\draw[black,fill=gray] (t3) ++(0,-0.30) coordinate(r1) rectangle +(1,1);
\draw[black,fill=lightgray] (r1) ++(-0.1,-0.1) coordinate(r2) rectangle +(1,1);
\draw[black,fill=gray] (r2) ++(-0.1,-0.1) coordinate(r3) rectangle +(1,1);
\draw[black,fill=lightgray] (r3) ++(-0.1,-0.1) coordinate(r4) rectangle +(1,1) node[pos=0.5]{$\Psi_j$};

\draw[->,thick] (t4) +(-2.4,0) -- +(-1.25,0);
\draw[->,thick] (t4) +(1.2,1.3) to[out=-90,in=40] +(0.3,0.9);
\draw (t4) node[cloud, draw, cloud puffs=12, cloud puff arc=120, aspect=1.5, inner ysep=1em,align=center] {\small{}Scattering};
\draw (t4) +(0,0.35) node{\scriptsize\subref{subfig:gscat}};
\draw (t4) +(0,-0.3) node{\tiny$\mathbf{x} \mapsto S\mathbf{x}$};
\draw[->,thick] (t4) +(1.25,0) -- +(2.2,0);

\draw[black,rounded corners] (t5) rectangle +(2.7,1.7) node[pos=0.5]{\parbox{2.5cm}{\centering\small Traditional Euclidean algorithms (e.g.,~SVM/PCA)}};

\end{tikzpicture}}}
\label{subfig:gsc}}
\vspace{-10pt}
\caption{Illustration of \subref{subfig:gscat} the proposed scattering feature extraction (see eqs.~\ref{eqn: zero layer},~\ref{eqn: first layer}, and~\ref{eqn: second layer}), and \subref{subfig:gsc} its application for graph data analysis.}
\label{fig:illustration}
\end{figure*}

The collection of graph scattering moments $S \xb = \{ S \xb (q), ~ S \xb (j, q), ~ S \xb (j, j', q) \}$ (illustrated in Fig.~\ref{subfig:gscat}) provides a rich set of multiscale invariants of the graph $G$. These can be used in supervised settings as input to graph classification or regression models, or in unsupervised settings to embed graphs into a Euclidean feature space for further exploration, as demonstrated in Sec.~\ref{sec: results}. 

\subsection{Stability and capacity of geometric scattering}
\label{sec: capacity}

In order to assess the utility of scattering features for representing graphs, two properties have to be considered: stability and capacity. First, the stability property aims to provide an upper bound on distances between similar graphs that only differ by types of deformations that can be treated as noise. This property has been the focus of both \citet{zou:graphCNNScat2018} and \citet{gama:diffScatGraphs2018}, and in particular the latter shows that a diffusion scattering transform yields features that are stable to graph structure deformations whose size can be computed via the diffusion framework~\citep{coifman:diffWavelets2006} that forms the basis for their construction. While there are some technical differences between the geometric scattering here and the diffusion scattering in \citet{gama:diffScatGraphs2018}, these constructions are sufficiently similar that we can expect both of them to have analogous stability properties. Therefore, we mainly focus here on the complementary property of the scattering transform capacity to provide a rich feature space for representing graph data without eliminating informative variance in them.

We note that even in the classical Euclidean case, while the stability of scattering transforms to deformations can be established analytically~\citep{mallat:scattering2012}, their capacity is typically examined by empirical evidence when applied to machine learning tasks~\citep[e.g.,][]{bruna:scatClass2011, sifre:rotoScatTexture2012,anden:deepScatSpectrum2014}. Similarly, in the graph processing settings, we examine the capacity of our proposed geometric scattering features via their discriminative power in graph data analysis tasks, which are described in detail in Sec.~\ref{sec: results}. We show that geometric scattering enables graph embedding in a relatively low dimensional Euclidean space, while preserving insightful properties in the data. Beyond establishing the capacity of our specific construction, these results also indicate the viability of graph scattering transforms as universal feature extractors on graph data, and complement the stability results established in \citet{zou:graphCNNScat2018} and \citet{gama:diffScatGraphs2018}. 

\subsection{Geometric scattering compared to other feed forward graph ConvNets}
\label{sec: comparison}

We give a brief comparison of geometric scattering with other graph ConvNets, with particular interest in isolating the key principles for building accurate graph ConvNet classifiers. We begin by remarking that like several other successful graph neural networks, the graph scattering transform is equivariant to vertex permutations (i.e., commutes with them) until the final features are extracted. This idea has been discussed in depth in various articles, including \citet{kondor:covariantCompNets2018}, so we limit the discussion to observing that the geometric scattering transform thus propagates nearly all of the information in $\xb$ through the multiple wavelet and absolute value layers, since only the absolute value operation removes information on $\xb$. As in \citet{verma2008:GCAPS-CNN}, we aggregate covariant responses via multiple summary statistics (i.e., moments), which are referred to there as a capsule. In the scattering context, at least, this idea is in fact not new and has been previously used in the Euclidean setting for the regression of quantum mechanical energies in~\citet{eickenberg:scatMoleculesJCP2018,eickenberg:3DSolidHarmonicScat2017} and texture synthesis in \citet{bruna:multiscaleMicrocanonical2018}. We also point out that, unlike many deep learning classifiers (graph included), a graph scattering transform extracts invariant statistics at each layer/order. These intermediate layer statistics, while necessarily losing some information in $\xb$ (and hence $G$), provide important coarse geometric invariants that eliminate needless complexity in subsequent classification or regression. Furthermore, such layer by layer statistics have proven useful in characterizing signals of other types \citep[e.g., texture synthesis in][]{NIPS2015_5633}.

A graph wavelet transform $\Psib^{(J)} \xb$ decomposes the geometry of $G$ through the lens of $\xb$, along different scales. Graph ConvNet algorithms also obtain multiscale representations of $G$, but several works, including \citet{atwood2016:DCNN} and \citet{zhang2018end11}, propagate information via a random walk. While random walk operators like $\Pb^t$ act at different scales on the graph $G$, per the analysis in Sec.~\ref{sec: graph wavelets} we see that $\Pb^t$ for any $t$ will be dominated by the low frequency responses of $\xb$. While subsequent nonlinearities may be able to recover this high frequency information, the resulting transform will most likely be unstable due to the suppression and then attempted recovery of the high frequency content of $\xb$. Alternatively, features derived from $\Pb^t \xb$ may lose the high frequency responses of $\xb$, which are useful in distinguishing similar graphs. The graph wavelet coefficients $\Psib^{(J)} \xb$, on the other hand, respond most strongly within bands of nearly non-overlapping frequencies, each with a center frequency $k_j$ that depends on $\Psib_j$. 

Finally, graph labels are often complex functions of both local and global subgraph structure within $G$. While graph ConvNets are adept at learning local structure within $G$, as detailed in \citet{verma2008:GCAPS-CNN} they require many layers to obtain features that aggregate macroscopic patterns in the graph. This is due to the use of fixed size filters, which often only incorporate information from the neighbors of a vertex. The training of such networks is difficult due to the limited size of many graph classification databases (see the supplementary information).
Geometric scattering transforms have two advantages in this regard: (a) the wavelet filters are designed; and (b) they are multiscale, thus incorporating macroscopic graph patterns in every layer/order. 

\vspace{-5pt}

\section{Application \& Results}
\label{sec: results}

To establish the geometric scattering features as an effective graph representation for data analysis, we examine their performance here in four graph data analysis applications. Namely, in Sec.~\ref{subsec: classification} we consider graph classification on social networks \citep[from][]{yanardag:DGK-SOCIAL}, in Sec.~\ref{subsec: low train} we consider the impact of low training data availability on classification, in Sec.~\ref{subsec: dim red} we examine dimensionality reduction aspects of geometric scattering, and finally, in Sec.~\ref{subsec: exploration} we consider data exploration of enzyme graphs, where geometric scattering enables unsupervised (descriptive) recovery of EC change preferences in enzyme evolution. A common theme in all these applications is the application of geometric scattering as an unsupervised task-independent feature extraction that embeds input graphs of varying sizes (with associated graph signals) into a Euclidean space formed by scattering features. Then, the extracted feature vectors are passed to traditional (Euclidean) machine learning algorithms, such as SVM for classification or PCA for dimensionality reduction, to perform downstream analysis. Our results show that our scattering features provide simplified representation (e.g., in dimensionality and extrapolation ability) of input graphs, which we conjecture is a result of their stability properties, while also being sufficiently rich to capture meaningful relations between graphs for predictive and descriptive purposes.

\subsection{Graph classification on social networks}
\label{subsec: classification}
As a first application of geometric scattering, we apply it to graph classification of social network data taken from \citet{yanardag:DGK-SOCIAL}. In particular, this work introduced six social network data sets extracted from scientific collaborations (COLLAB), movie collaborations (IMDB-B \& IMDB-M), and Reddit discussion threads (REDDIT-B, REDDIT-5K, REDDIT-12K). There are also biochemistry data sets often used in the graph classification literature; for completeness, we include in the supplemental materials further results on these data sets. A brief description of each data set can also be found in the supplement.

The social network data provided by \citet{yanardag:DGK-SOCIAL} contains graph structures but no associated graph signals. Therefore we compute the eccentricity (for connected graphs) and clustering coefficient of each vertex, and use these as input signals to the geometric scattering transform. In principle, any general node characteristic could be used, although we remark that $\xb = \db$, the vertex degree vector, is not useful in our construction since $\Psib_j \db = \zerob$. After computing the scattering moments\footnote{We use the normalized scattering moments for classification, since they perform slightly better than the un-normalized moments. Also we use $J=5$ and $q=4$ for all scattering feature generations.} of these two input signals, they are concatenated to form a single vector. This scattering feature vector is a consistent Euclidean representation of the graph, which is independent of the original graph sizes (i.e., number of vertices or edges), and thus we can apply any traditional classifier to it. In particular, we use here the standard SVM classifier with an RBF kernel, which is popular and effective in many applications and also performs well in this case.  

\begin{table*}[!htb]
\centering
\tiny
\caption{Comparison of the proposed GS-SVM classifier with leading graph kernel and deep learning methods on social graph datasets.}
\label{tbl:results}
\adjustbox{width=0.8\linewidth}{\begin{tabular}{r|c|c|c|c|c|c|l}
\hhline{~------~}
~ & COLLAB & IMDB-B & IMDB-M & REDDIT-B & REDDIT-5K & REDDIT-12K & \\
\hhline{~------~}\hhline{~------~}
WL & $77.82 \pm 1.45$ & $71.60 \pm 5.16$ & N/A & $78.52 \pm 2.01$ & $50.77 \pm 2.02$ & $34.57 \pm 1.32$ & \multirow{5}{*}{\hspace{-5pt}\rotatebox[origin=c]{-90}{$\overbrace{\hspace{35pt}}^{\text{Graph kernel}}$}}\\
Graphlet & $73.42 \pm 2.43$ & $65.40 \pm 5.95$ & N/A & $77.26 \pm 2.34$ & $39.75 \pm 1.36$ & $25.98 \pm 1.29$&\\
WL-OA & $80.70 \pm 0.10$ & N/A & N/A & $89.30 \pm 0.30$ & N/A & N/A & \\
GK & $72.84 \pm 0.28$ & $65.87 \pm 0.98$ & $43.89 \pm 0.38$ & $77.34 \pm 0.18$ & $41.01 \pm 0.17$ & N/A & \\
DGK & $73.00 \pm 0.20$ & $66.90 \pm 0.50$ & $44.50 \pm 0.50$ & $78.00 \pm 0.30$ & $41.20 \pm 0.10$ & $32.20 \pm 0.10$ &\\
\hhline{~------~}\hhline{~------~}
DGCNN & $73.76 \pm 0.49$ & $70.03 \pm 0.86$ & $47.83 \pm 0.85$ & N/A & $48.70 \pm 4.54$ & N/A &\multirow{6}{*}{\hspace{-5pt}\rotatebox[origin=c]{-90}{$\overbrace{\hspace{40pt}}^{\text{Deep learning}}$}}\\
2D CNN & $71.33 \pm 1.96$ & $70.40 \pm 3.85$ & N/A & $89.12 \pm 1.70$ & $52.21 \pm 2.44$ & $48.13 \pm 1.47$ &\\
PSCN ($k=10$) & $72.60 \pm 2.15$ & $71.00 \pm 2.29$ & $45.23 \pm 2.84$ & $86.30 \pm 1.58$ & $49.10 \pm 0.70$ & $41.32 \pm 0.42$ &\\
GCAPS-CNN & $77.71 \pm 2.51$ & $71.69 \pm 3.40$ & $48.50 \pm 4.10$ & $87.61 \pm 2.51$ & $50.10 \pm 1.72$ & N/A &\\
S2S-P2P-NN & $81.75 \pm 0.80$ & $73.80 \pm 0.70$ & $51.19 \pm 0.50$ & $86.50 \pm 0.80$ & $52.28 \pm 0.50$ & $42.47 \pm 0.10$ & \\
GIN-0 (MLP-SUM) & $80.20 \pm 1.90$ & $75.10 \pm 5.10$ & $52.30 \pm 2.80$ & $92.40 \pm 2.50$ & $57.50 \pm1.50$ & N/A \\
\hhline{~------~}\hhline{~------~}
\emph{GS-SVM} & $79.94 \pm 1.61$ & $71.20 \pm 3.25$ & $48.73 \pm 2.32$ & $89.65 \pm 1.94$ & $53.33 \pm 1.37$ & $45.23 \pm 1.25$ &\\
\hhline{~------~}
\end{tabular}}
\end{table*}

We evaluate the classification results of our SVM-based geometric scattering classification (GS-SVM) using ten-fold cross validation (as explained, for completeness, in the supplament), which is standard practice in other graph classification works. We compare our results to $11$ prominent methods that report results for most, if not all, of the considered datasets. Out of these, five are graph kernel methods, namely: Weisfeiler-Lehman graph kernels~\citep[WL,][]{shervashidze2011:WL}, propagation kernel~\citep[PK,][]{neumann2012:PK}, Graphlet kernels~\citep{shervashidze2009:Graphlet}, Random walks~\citep[RW,][]{gartner2003:RW}, deep graph kernels~\citep[DGK,][]{yanardag:DGK-SOCIAL}, and Weisfeiler-Lehman optimal assignment kernels~\citep[WL-OA,][]{NIPS2016_6166}. The other six are recent geometric deep learning algorithms: deep graph convolutional neural network~\citep[DGCNN,][]{zhang2018end11}, Graph2vec~\citep{narayanan2017:graph2vec}, 2D convolutional neural networks~\citep[2DCNN,][]{tixier2017:2DCNN}, covariant compositional networks~\citep[CCN,][]{kondor2018:CCN}, Patchy-san~\citep[PSCN,][with $k=10$]{niepert2016:PSCN}, diffusion convolutional neural networks~\citep[DCNN,][]{atwood2016:DCNN}, graph capsule convolutional neural networks~\citep[GCAPS-CNN,][]{verma2008:GCAPS-CNN}, recurrent neural network autoencoders~\citep[S2S-N2N-PP,][]{Taheri:graphRNN2018}, and the graph isomorphism network~\citep[GIN,][]{xu2018how}.

Following the standard format of reported classification performances for these methods (per their respective references, see also the supplement), our results are reported in the form of average accuracy $\pm$ standard deviation (in percentages) over the ten cross-validation folds. We note that since some methods are not reported for all datasets, we mark N/A when appropriate. Table~\ref{tbl:results} reports the results.

The geometric scattering transform and related variants presented in \citet{zou:graphCNNScat2018} and \citet{gama:diffScatGraphs2018} is a mathematical model for graph ConvNets. However, it is natural to ask if this model accurately reflects what is done in practice. A useful model may not obtain state of the art performance, but should be competitive with the current state of the art, lest the model may not capture the underlying complexity of the most powerful methods. Examining Table~\ref{tbl:results} one can see that the GS-SVM classifier matches or outperforms all but the two most recent methods, i.e., S2S-N2N-PP~\citep{Taheri:graphRNN2018} and GIN~\citep{xu2018how}. With regards to these two approaches, the GS-SVM outperforms S2S-N2N-PP~\citep{Taheri:graphRNN2018} on $\nicefrac{3}{6}$ datasets. Finally, while GIN~\citep{xu2018how} outperforms geometric scattering on $\nicefrac{5}{6}$ datasets, the results on COLLAB and IMDB-B are not statistically significant, and on the REDDIT datasets the geometric scattering approach trails only GIN~\citep{xu2018how}. We thus conclude that the geometric scattering transform yields a rich set of invariant statistical moments, which have nearly the same capacity as the current state of the art in graph neural networks.

\subsection{Classification with low training-data availability} 
\label{subsec: low train}
Many modern deep learning methods require large amounts of training data to generate representative features. On the contrary, geometric scattering features are based on each graph without any training processes. In this section, we demonstrate the performance of the GS-SVM under low training-data availability and show that the scattering features can embed enough graph information that even under extreme conditions (e.g. only 20\% training data), they can still maintain relatively good classification results.

\begin{figure}[!b]
    \centering
    \adjustbox{width=0.99\linewidth}{\begin{tikzpicture}
    \node(0,0){\includegraphics[scale=0.5,trim=0.4in 1.48in 0.29in 0.55in,clip]{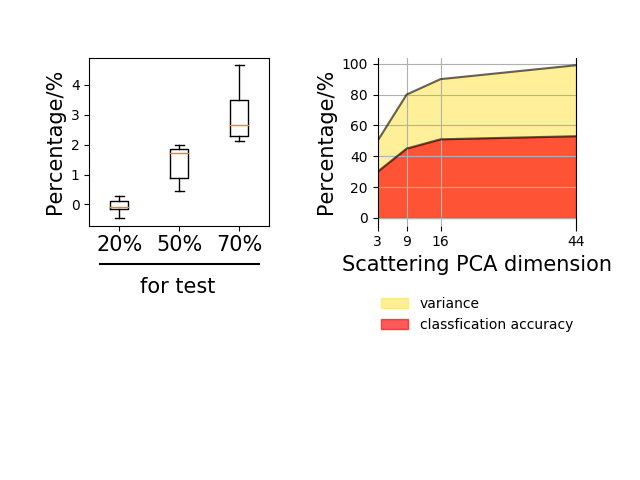}};
    \draw(-2.8,-1.4) node{\subfigure[]{\hspace{20pt}\label{subfig:small-training}}};
    \draw(0.45,-1.4) node{\subfigure[]{\hspace{20pt}\label{subfig:dim-reduction}}};
    \end{tikzpicture}}
    \caption{\subref{subfig:small-training} Box plot showing the drop in SVM classification accuracy over social graph datasets when reducing training set size (horizontal axis marks portion of data used for testing); \subref{subfig:dim-reduction} Relation between explained variance, SVM classification accuracy, and PCA dimensions over scattering features in ENZYMES dataset.}
    \label{fig:enzyme-PCA}
\end{figure}

We performed graph classification under four training/validation/test splits: 80\%/10\%/10\%, 70\%/10\%/20\%, 40\%/10\%/50\% and 20\%/10\%/70\%. We did 10-fold, 5-fold and 2-fold cross validation for the first three splits. For the last split, we randomly formed a 10 folds pool, from which we randomly selected 3 folds for training/validation and repeated this process ten times. Detailed classification results can be found in the supplement. Following Sec.~\ref{subsec: classification}, we discuss the classification accuracy on six social datasets under these splits. When the training data is reduced from 90\% to 80\%, the classification accuracy in fact increased by 0.047\%, which shows the GS-SVM classification accuracy is not affected by the decrease in training size. Further reducing the training size to 50\% results in an average decrease of classification accuracy of 1.40\% while from 90\% to 20\% causes an average decrease of 3.00\%. Fig.~\ref{fig:enzyme-PCA} gives a more nuanced statistical description of these results.

\subsection{Dimensionality reduction} \label{subsec: dim red}
We now consider the viability of scattering-based embedding for dimensionality reduction of graph data. As a representative example, we consider here the ENZYMES dataset introduced in \citet{borgwardt:PROTEINS}, which contains $600$ enzymes evenly split into six enzyme classes (i.e., $100$ enzymes from each class). While the Euclidean notion of dimensionality is not naturally available in graph data, we note that graphs in this dataset have, on average, 124.2 edges, 29.8 vertices, and 3 features per vertex. Therefore, the data here can be considered significantly high dimensional in its original representation, which is not amenable to traditional dimensionality reduction techniques. 

To perform scattering-based dimensionality reduction, we applied PCA to geometric scattering features extracted from input enzyme graphs in the data, while choosing the number of principal components to capture 99\%, 90\%, 80\% and 50\% explained variance. For each of these thresholds, we computed the mean classification accuracy (with ten-fold cross validation) of SVM applied to the GS-PCA low dimensional space, as well as the dimensionality of this space. The relation between dimensionality, explained variance, and SVM accuracy is shown in Fig.~\ref{fig:enzyme-PCA}, where we can observe that indeed geometric scattering combined with PCA enables significant dimensionality reduction (e.g., to $\mathbb{R}^{16}$ with 90\% exp.\ variance) with only a small impact on classification accuracy. Finally, we also consider the PCA dimensionality of each individual enzyme class in the data (in the scattering feature space), as we expect scattering to reduce the variability in each class w.r.t.\ the full feature space. Indeed, in this case, individual classes have 90\% exp.\ variance PCA dimensionality ranging between 6 and 10, which is significantly lower than the 16 dimensions of the entire PCA space. We note that similar results can also be observed for the social network data discussed in previous sections, where on average 90\% explained variances are captured by nine dimensions, yielding a drop of 3.81\% in mean SVM accuracy; see the supplement for complete results.

\subsection{Data exploration: Enzyme class exchange preferences}
\label{subsec: exploration}

Geometric scattering essentially provides a task independent representation of graphs in a Euclidean feature space. Therefore, it is not limited to supervised learning applications, and can be also utilized for exploratory graph-data analysis, as we demonstrate in this section. We focus our discussion in particular on the ENZYMES dataset described in the previous section. Here, geometric scattering features can be considered as providing ``signature'' vectors for individual enzymes, which can be used to explore interactions between the six top level enzyme classes, labelled by their Enzyme Commission (EC) numbers~\citep{borgwardt:ENZYMES}. In order to emphasize the properties of scattering-based feature extraction, rather than downstream processing, we mostly limit our analysis of the scattering feature space to linear operations such as principal component analysis (PCA).

\begin{figure}[!b]
    \centering
    \subfigure[Observed]{\includegraphics[width=0.49\linewidth]{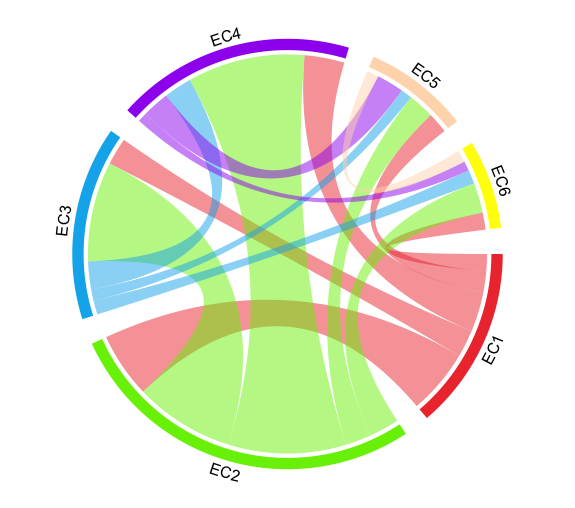}\label{subfig:observed-EC}}\hfill
    \subfigure[Inferred]{\includegraphics[width=0.49\linewidth]{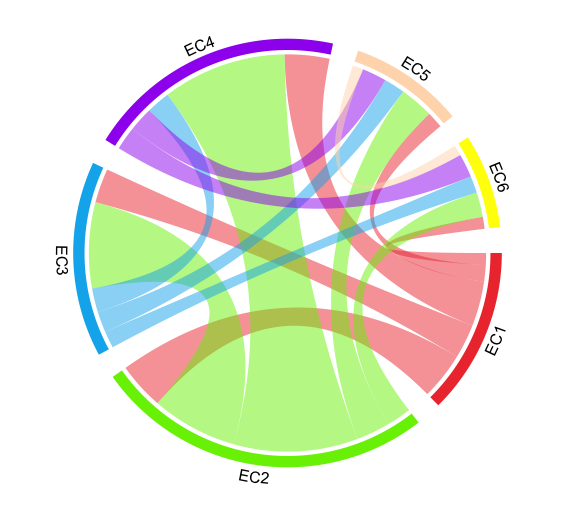}\label{subfig:inferred-EC}}
    \caption{Comparison of EC exchange preferences in enzyme evolution: \subref{subfig:observed-EC} observed in~\citet{cuesta:EC-evolution}, and \subref{subfig:inferred-EC} inferred from scattering features via $\text{pref}(\text{EC-}i,\text{EC-}j) := w_j\,\cdot\,\left[\min\left\{\frac{D(i,j)}{D(i,i)},\frac{D(j,i)}{D(j,j)}\right\}\right]^{-1}$; $w_j\,=\,$ portion of enzymes in EC-$j$ that choose another EC as their nearest subspace; $D(i,j)\,{=}\,$mean dist.\ of enzymes in EC-$i$ from PCA ($90\%$ exp.\ var.) subspace of EC-$j\,$.
    Our inference \subref{subfig:inferred-EC} mainly recovers \subref{subfig:observed-EC}.}
    \label{fig:ec-pref}
\end{figure}

To explore the scattering feature space, and the richness of information captured by it, we use it to infer relations between EC classes. First, for each enzyme $e$, with scattering feature vector $\vb_e$ (i.e., with $S\xb$ for all vertex features~$\xb$), we compute its distance from class EC-$j$, with PCA subspace $\mathcal{C}_j$, as the projection distance: $\text{dist}(e,\text{EC-}j) = \|\vb_e - \text{proj}_{\mathcal{S}_j} \vb_e \|$. Then, for each enzyme class EC-$i$, we compute the mean distance of enzymes in it from the subspace of each EC-$j$ class as $D(i,j) = \text{mean}\{\text{dist}(e,\text{EC-}j) : e \in \text{EC-}i\}$. These distances are summarized in the supplement, as well as the proportion of points from each class that have their true EC as their nearest (or second nearest) subspace in the scattering feature space. In general, $48\%$ of enzymes select their true EC as the nearest subspace (with additional $19\%$ as second nearest), but these proportions vary between individual EC classes. Finally, we use these scattering-based distances to infer EC exchange preferences during enzyme evolution, which are presented in Fig.~\ref{fig:ec-pref} and validated with respect to established preferences observed and reported in~\citet{cuesta:EC-evolution}. We note that the result there is observed independently from the ENZYMES dataset. In particular, the portion of enzymes considered from each EC is different between these data, since \citet{borgwardt:PROTEINS} took special care to ensure each EC class in ENZYMES has exactly 100 enzymes in it. However, we notice that in fact the portion of enzymes (in each EC) that choose the wrong EC as their nearest subspace, which can be considered as EC ``incoherence'' in the scattering feature space, correlates well with the proportion of evolutionary exchanges generally observed for each EC in~\citet{cuesta:EC-evolution}, and therefore we use these as EC weights (see Fig.~\ref{fig:ec-pref}). Our results in Fig.~\ref{fig:ec-pref} demonstrate that scattering features are sufficiently rich to capture relations between enzyme classes, and indicate that geometric scattering has the capacity to uncover descriptive and exploratory insights in graph data analysis.

\section{Conclusion} \label{sec: conclusion}

We presented the geometric scattering transform as a deep filter bank for feature extraction on graphs, which generalizes the Euclidean scattering transform. A reasonable criticism of the scattering theory approach to understanding geometric deep learning is that it is not clear if the scattering model is a suitable facsimile for powerful graph neural networks that are obtaining impressive results on graph classification tasks and related graph data analysis problems. In this paper we showed that in fact, at least empirically, this line of criticism is unfounded and indeed further theoretical study of geometric scattering transforms on graphs is warranted. Our evaluation results on graph classification and data exploration show the potential of the produced scattering features to serve as universal representations of graphs. Indeed, classification using these features with relatively simple classifier models, dimension reduced feature sets, and small training sets nevertheless reach high accuracy results on most commonly used graph classification datasets. Finally, the geometric scattering features provide a new way for computing and considering global graph representations, independent of specific learning tasks. They raise the possibility of embedding entire graphs in Euclidean space and computing meaningful distances between graphs, which can be used for both supervised and unsupervised learning, as well as exploratory analysis of graph-structured data.

\onecolumn

\subsubsection*{Acknowledgments}
F.G. is supported by the grant P42 ES004911 through the National Institute of Environmental Health Sciences of the National Institutes of Health.
M.H. is supported by the Alfred P. Sloan Fellowship (grant FG-2016-6607), the DARPA YFA (grant D16AP00117), and NSF grant 1620216.

\bibliographystyle{icml2019}
\bibliography{MainBib,comparisons}

\newpage
\small 
\begin{appendices}

\section{Detailed graph classification comparison}
\label{apx: results}

\begin{table}[!htb]
\caption{Comparison of the proposed graph scattering classifier (GSC) with graph kernel methods and deep learning methods on biochemistry \& social graph datasets. (Remark\footnotemark[1]: DCNN using different training/test split)}
\centering
\scalebox{0.9}{\begin{sideways}
\begin{tabular}{r|c|c|c|c|c|c|c|l}
\hhline{~-------~}
~ & NCI1 & NCI109 & D\&D & PROTEINS & MUTAG & PTC & ENZYMES &\\
\hhline{~-------~}\hhline{~-------~}
WL & $84.46 \pm 0.45$ & $85.12 \pm 0.29$ & $78.34 \pm 0.62$ & $72.92 \pm 0.56$ & $84.11 \pm 1.91$ & $59.97 \pm 1.60$ & $55.22 \pm 1.26$ & \multirow{6}{*}{\hspace{-5pt}\rotatebox[origin=c]{-90}{$\overbrace{\hspace{65pt}}^{\text{Graph kernel}}$}}\\
PK & $82.54 \pm 0.47$ & N/A & $78.25 \pm 0.51$ & $73.68 \pm 0.68$ & $76.00 \pm 2.69$ & $59.50 \pm 2.44$ & N/A & \\
Graphlet & $70.5 \pm 0.2$ & $69.3 \pm 0.2$ & $79.7 \pm 0.7$ & $72.7 \pm 0.6$ & $85.2 \pm 0.9$ & $54.7 \pm 2.0$ & $30.6 \pm 1.2$ & \\
WL-OA & $86.1 \pm 0.2$ & $86.3 \pm 0.2$ & $79.2 \pm 0.4$ & $76.4 \pm 0.4$ & $84.5 \pm 1.7$ & $63.6 \pm 1.5$ & $59.9 \pm 1.1$ &\\
GK & $62.28 \pm 0.29$ & $62.60 \pm 0.19$ & $78.45 \pm 0.26$ & $71.67 \pm 0.55$ & $81.39 \pm 1.74$ & $57.26 \pm 1.41$ & $26.61 \pm 0.99$ & \\
DGK & $80.3 \pm 0.4$ & $80.3 \pm 0.3$ & $73.09 \pm 0.25$ & $75.7 \pm 0.50$ & $87.4 \pm 2.7$ & $60.1 \pm 2.5$ & $53.4 \pm 0.9$ & \\
\hhline{~-------~}\hhline{~-------~}
DGCNN & $74.44 \pm 0.47$ & N/A & $79.37 \pm 0.94$ & $75.54 \pm 0.94$ & $85.83 \pm 1.66$ & $58.59 \pm 2.47$ & $51.00 \pm 7.29$ & \multirow{9}{*}{\hspace{-5pt}\rotatebox[origin=c]{-90}{$\overbrace{\hspace{102pt}}^{\text{Deep learning}}$}}\\
graph2vec & $73.22 \pm 1.81$ & $74.26 \pm 1.47$ & N/A & $73.30 \pm 2.05$ & $83.15 \pm 9.25$ & $60.17 \pm 6.86$ & N/A & \\
2D CNN & N/A & N/A & N/A & $77.12 \pm 2.79$ & N/A & N/A & N/A & \\
CCN & $76.27 \pm 4.13$ & $75.54 \pm 3.36$ & N/A & N/A & $91.64 \pm 7.24$ & $70.62 \pm 7.04$ & N/A & \\
PSCN ($k=10$) & $76.34 \pm 1.68$ & N/A & $76.27 \pm 2.15$ & $75.00 \pm 2.51$ & $88.95 \pm 4.37$ & $62.29 \pm 5.68$ & N/A & \\
DCNN & $56.61 \pm 1.04$ & $57.47 \pm 1.22$ & $58.09 \pm 0.53$ & $61.29 \pm 1.60$ & $56.60 \pm 2.89$ & $56$\footnotemark[1] & $42.44 \pm 1.76$ & \\
GCAPS-CNN & $82.72 \pm 2.38$ & $81.12 \pm 1.28$ & $77.62 \pm 4.99$ & $76.40 \pm 4.17$ & N/A & $66.01 \pm 5.91$ & $61.83 \pm 5.39$ & \\
S2S-P2P-NN & $83.72 \pm 0.4$ & $83.64 \pm 0.3$ & N/A & $76.61 \pm 0.5$ & $89.86 \pm 1.1$ & $64.54 \pm 1.1$ & $63.96 \pm 0.6$ & \\
GIN-0 (MLP-SUM) & $82.70 \pm 1.60$ & N/A & N/A & $76.20 \pm 2.80$ & $89.40 \pm 5.60$ & $64.60 \pm 7.00$ & N/A & \\
\hhline{~-------~}\hhline{~-------~}
\emph{GS-SVM} & $79.14 \pm 1.28$ & $77.95 \pm 1.25$ & $75.04 \pm 3.64$ & $74.11 \pm 4.02$ & $83.57 \pm 6.75$ & $63.94 \pm 7.38$ & $56.83 \pm 4.97$ & \\

\hhline{~-------~}
\end{tabular}
\end{sideways} \qquad \begin{sideways}
\begin{tabular}{r|c|c|c|c|c|c|l}
\hhline{~------~}
~ & COLLAB & IMDB-B & IMDB-M & REDDIT-B & REDDIT-5K & REDDIT-12K \\
\hhline{~------~}\hhline{~------~}
WL & $77.82 \pm 1.45$ & $71.60 \pm 5.16$ & N/A & $78.52 \pm 2.01$ & $50.77 \pm 2.02$ & $34.57 \pm 1.32$ & \multirow{6}{*}{\hspace{-5pt}\rotatebox[origin=c]{-90}{$\overbrace{\hspace{65pt}}^{\text{Graph kernel}}$}}\\
PK & N/A & N/A & N/A & N/A & N/A & N/A & \\
Graphlet & $73.42 \pm 2.43$ & $65.4 \pm 5.95$ & N/A & $77.26 \pm 2.34$ & $39.75 \pm 1.36$ & $25.98 \pm 1.29$ &\\
WL-OA & $80.7 \pm 0.1$ & N/A & N/A & $89.3 \pm 0.3$ & N/A & N/A &\\
GK & $72.84 \pm 0.28$ & $65.87 \pm 0.98$ & $43.89 \pm 0.38$ & $77.34 \pm 0.18$ & $41.01 \pm 0.17$ & N/A & \\
DGK & $73.0 \pm 0.2$ & $66.9 \pm 0.5$ & $44.5 \pm 0.5$ & $78.0 \pm 0.3$ & $41.2 \pm 0.1$ & $32.2 \pm 0.1$ & \\
\hhline{~------~}\hhline{~------~}
DGCNN & $73.76 \pm 0.49$ & $70.03 \pm 0.86$ & $47.83 \pm 0.85$ & N/A & $48.70 \pm 4.54$ & N/A &  \multirow{9}{*}{\hspace{-5pt}\rotatebox[origin=c]{-90}{$\overbrace{\hspace{102pt}}^{\text{Deep learning}}$}}\\
graph2vec & N/A & N/A & N/A & N/A & N/A & N/A & \\
2D CNN & $71.33 \pm 1.96$ & $70.40 \pm 3.85$ & N/A & $89.12 \pm 1.7$ & $52.21 \pm 2.44$ & $48.13 \pm 1.47$ & \\
CCN & N/A & N/A & N/A & N/A & N/A & N/A & \\
PSCN ($k=10$) & $72.60 \pm 2.15$ & $71.00 \pm 2.29$ & $45.23 \pm 2.84$ & $86.30 \pm 1.58$ & $49.10 \pm 0.7$ & $41.32 \pm 0.42$ & \\
DCNN & $52.11 \pm 0.71$ & $49.06 \pm 1.37$ & $33.49 \pm 1.42$ & N/A & N/A & N/A & \\
GCAPS-CNN & $77.71 \pm 2.51$ & $71.69 \pm 3.40$ & $48.50 \pm 4.1$ & $87.61 \pm 2.51$ & $50.10 \pm 1.72$ & N/A & \\
S2S-P2P-NN & $81.75 \pm 0.8$ & $73.8 \pm 0.7$ & $51.19 \pm 0.5$ & $86.50 \pm 0.8$ & $52.28 \pm 0.5$ & $42.47 \pm 0.1$ & \\
GIN-0 (MLP-SUM) & $80.20 \pm 1.90$ & $75.10 \pm 5.10$ & $52.30 \pm 2.80$ & $92.40 \pm 2.50$ & $57.50 \pm 1.50$ & N/A \\
\hhline{~------~}\hhline{~------~}
\emph{GS-SVM} & $79.94 \pm 1.61$ & $71.20 \pm 3.25$ & $48.73 \pm 2.32$ & $89.65 \pm 1.94$ & $53.33 \pm 1.37$ & $45.23 \pm 1.25$ & \\
\hhline{~------~}
\end{tabular}\end{sideways}}
\end{table}

All results come from the respective papers that introduced the methods, with the exception of: \begin{enumerate*} \item[(1)] social network results of WL, from~\citet{tixier2017:2DCNN}; \item[(2)] biochemistry and social results of DCNN, from~\citet{verma2008:GCAPS-CNN}; \item[(3)] biochemistry, except for D\&D, and social result of GK, from~\citet{yanardag:DGK-SOCIAL}; \item[(4)] D\&D of GK is from~\citet{niepert2016:PSCN}; and \item[(5)] for Graphlets, biochemistry results from~\citet{NIPS2016_6166}, social results from~\citet{tixier2017:2DCNN}. \end{enumerate*}

\section{Detailed tables for scattering feature space analysis from Section~4}
\label{apx: exploration}

\begin{table}[H]
    \centering
    \caption{Classification accuracy with different training/validaion/test splits over scattering features (unnorm. moments)}
    \begin{tabular}{|c|c|c|c|c|}
        \hline
        \multirow{2}{*}{Dataset} & \multicolumn{4}{c|}{SVM accuracy} \\
        \hhline{~----}
        & 80\%/10\%/10\% & 70\%/10\%/20\% & 40\%/10\%/50\% & 20\%/10\%/70\% \\
        \hline\hline
        NCI1 & $79.80 \pm 2.24$ & $78.13 \pm 2.07$ & $ 76.37 \pm 0.27$ & $73.60 \pm 0.68$\\
        \hline
        NCI109 & $77.66 \pm 1.78$ & $77.54 \pm 1.44$ & $74.41 \pm 0.14$ &$ 72.36 \pm 0.74$\\
        \hline
        D\&D & $76.57 \pm 3.76$ & $76.74 \pm 2.32$ & $ 76.32\pm 0.59 $ & $ 75.58\pm 0.81$ \\
        \hline
        PROTEINS & $74.03 \pm 4.20$ & $74.30 \pm 2.49$ & $ 73.32 \pm 1.68$ & $ 73.01\pm 1.94$\\
        \hline
        MUTAG & $84.04 \pm 6.71$ & $82.99 \pm 6.97$ & $78.72 \pm 3.19$ & $77.47 \pm 4.41$\\
        \hline
        PTC & $66.32 \pm 7.54$ & $64.83 \pm 2.13$ & $61.92 \pm 1.45$ & $56.75 \pm 2.88$\\
        \hline
        ENZYMES & $53.83 \pm 6.71$ & $52.50 \pm 5.35$ & $44.50 \pm 3.83$ & $36.38 \pm 1.93$\\
        \hline
        COLLAB & $76.88 \pm 1.13$ & $76.98 \pm 0.97$ & $76.42 \pm 0.82$ & $74.63 \pm 1.05$\\
        \hline
        IMDB-B & $70.80 \pm 3.54$ & $70.60 \pm 2.85$ & $69.10 \pm 1.90$ & $67.81 \pm 0.98$\\
        \hline
        IMDB-M & $48.93 \pm 4.77$ & $49.00 \pm 1.97$ & $47.20 \pm 1.47$ & $44.28 \pm 1.87$\\
        \hline
        REDDIT-B & $88.30 \pm 2.08$ & $88.75 \pm 0.96$ & $86.40 \pm 0.40$ & $86.18 \pm 0.32$\\
        \hline
        REDDIT-5K & $50.71 \pm 2.27$ & $50.87 \pm 1.37$ & $50.10 \pm 0.41$ & $48.37 \pm 0.76$\\
        \hline
        REDDIT-12K & $41.35 \pm 1.05$ & $41.05 \pm 0.70$ & $39.36 \pm 1.30$ & $37.71 \pm 0.42$\\
        \hline
    \end{tabular}
    \label{tab:train-size}
\end{table}

\begin{table}[H]
    \centering
    \caption{Classification accuracy and dimensionality reduction with PCA over scattering features (unnorm. moments)}
    \begin{tabular}{|c|c|c|c|c|c|c|c|c|}
        \hline
        \multirow{2}{*}{Dataset} & \multicolumn{4}{c|}{SVM accuracy w.r.t variance covered} & \multicolumn{4}{c|}{PCA dimensions w.r.t variance covered} \\
        \hhline{~--------}
        & 50\% & 80\% & 90\% & 99\%& 50\% & 80\%& 90\%&99\%\\
        \hline\hline
        NCI1 & $72.41 \pm 2.36$ & $73.89 \pm 2.57$ & $73.89 \pm 1.33$ & $78.22 \pm 1.95$ & 18 & 32 & 43 & 117\\
        \hline
        NCI109 & $70.85 \pm 2.59$ & $71.84 \pm 2.38$ & $72.33 \pm 2.24$ & $76.69 \pm 1.02$ & 19 & 32 & 43 & 114\\
        \hline
        D\&D & $75.21 \pm 3.17$ & $75.13 \pm 3.68$ & $74.87 \pm 3.99$ & $76.92 \pm 3.37$ & 10 & 35 & 44 & 122\\
        \hline
        PROTEINS & $70.80 \pm 3.43$ & $74.20 \pm 3.06$ & $74.67 \pm 3.33$ & $74.57 \pm 3.42$ & 2 & 5 & 10 & 36\\
        \hline
        MUTAG & $77.51 \pm 10.42$ & $80.32 \pm 8.16$ & $82.40 \pm 10.92$ & $84.09 \pm 9.09$ & 4 & 8 & 13 & 34\\
        \hline
        PTC & $58.17 \pm 8.91$ & $60.50 \pm 9.96$ & $58.70 \pm 6.93$ & $63.68 \pm 3.97$ & 7 & 14 & 21 & 62\\
        \hline
        ENZYMES & $29.67 \pm 4.46$ & $45.33 \pm 6.62$ & $50.67 \pm 5.44$ & $52.50 \pm 8.89$ & 3 & 9 & 16 & 44\\
        \hline
        COLLAB & $62.86 \pm 1.36$ & $71.68 \pm 2.06$ & $73.22 \pm 2.29$ & $76.54 \pm 1.41$ & 2 & 6 & 9 & 32\\
        \hline
        IMDB-B & $58.30 \pm 3.44$ & $66.10 \pm 3.14$ & $68.80 \pm 4.31$ & $68.40 \pm 4.31$ & 2 & 4 & 8 & 24\\
        \hline
        IMDB-M & $41.00 \pm 4.86$ & $46.40 \pm 4.48$ & $45.93 \pm 3.86$ & $48.27 \pm 3.23$ & 2 & 5 & 8 & 20\\
        \hline
        REDDIT-B & $71.05 \pm 2.39$ & $78.95 \pm 2.42$ & $83.75 \pm 1.83$ & $86.95 \pm 1.78$ & 2 & 5 & 8 & 24\\
        \hline
        REDDIT-5K & $40.97 \pm 2.06$ & $45.71 \pm 2.21$ & $47.43 \pm 1.90$ & $49.65 \pm 1.86$ & 2 & 6 & 10 & 27\\
        \hline
        REDDIT-12K & $28.22 \pm 1.64$ & $33.36 \pm 0.93$ & $34.71 \pm 1.52$ & $38.39 \pm 1.54$ & 2 & 5 & 9 & 27\\
        \hline
    \end{tabular}
    \label{tab:dimensions-cls}
\end{table}

\begin{table}[H]
    \centering
    \caption{Dimensionality reduction with PCA over scattering features (unnorm. moments)}
    \begin{tabular}{|c|c|c|c|c|c|c|c|c|c|}
        \hline
        \multirow{2}{*}{Dataset} & \multicolumn{2}{c|}{SVM accuracy} & \multicolumn{7}{c|}{PCA dimensions ($>90\%$ variance)} \\
        \hhline{~---------}
        & PCA & Full & All classes & \multicolumn{6}{c|}{Per class} \\
        \hline\hline
        ENZYMES & $50.67 \pm 5.44$ & $53.83 \pm 6.71$ & 16 & 9 & 8 & 8 & 9 & 10 & 6 \\
        \hline
    \end{tabular}
    \label{tab:dimensions}
\end{table}

\begin{table}[H]
    \centering
    \caption{EC subspace analysis in scattering feature space of ENZYMES~\citep{borgwardt:ENZYMES}}
    \begin{tabular}{|c|rrrrrr|rrr|}
        \hline
        \multirow{3}{*}{\parbox{33pt}{\centering{}Enzyme\\Class:}} & \multicolumn{6}{c|}{Mean distance to subspace of class} & \multicolumn{3}{c|}{True class as} \\
        & EC-1 & EC-2 & EC-3 & EC-4 & EC-5 & EC-6 & $1^\text{st}$ & $\quad 2^\text{nd}$ & $3^\text{rd}$-$6^\text{th}$ \\
        & \multicolumn{6}{c|}{measured via PCA projection/reconstruction distance} & \multicolumn{3}{c|}{nearest subspace} \\
        \hline
        EC-1 & 18.15 & 98.44 & 75.47 & 62.87 & 53.07 & 84.86 & 45\% & 28\% & 27\%\\
        EC-2 & 22.65 & 9.43 & 30.14 & 22.66 & 18.45 & 22.75 & 53\% & 24\% & 23\%\\
        EC-3 & 107.23 & 252.31 & 30.4 & 144.08 & 117.24 & 168.56 & 32\% & 7\% & 61\%\\
        EC-4 & 117.68 & 127.27 & 122.3 & 29.59 & 94.3 & 49.14 & 24\% & 12\% & 64\%\\
        EC-5 & 45.46 & 66.57 & 60 & 50.07 & 15.09 & 58.22 & 67\% & 21\% & 12\%\\
        EC-6 & 62.38 & 58.88 & 73.96 & 51.94 & 59.23 & 13.56 & 67\% & 21\% & 12\%\\
        \hline
    \end{tabular}
    \label{tab:enzymes}
\end{table}

\section{Detailed Dataset Descriptions}
\label{apx: datasets}

The details of the datasets used in this work are as follows:
\begin{description}

\item[NCI1]\citep{wale:NC1-NCI109} contains 4,110 chemical compounds as graphs, with 37 node features. Each compound is labeled according to is activity against non-small cell lung cancer and ovarian cancer cell lines, and these labels serve as classification goal on this data.

\item[NCI109] \citep{wale:NC1-NCI109} is similar to NCI1, but with 4,127 chemical compounds and 38 node features. 

\item[MUTAG] \citep{debnath:MUTAG} consists of 188 mutagenic aromatic and heteroaromatic nitro compounds (as graphs) with 7 node features. The classification here is binary (i.e., two classes), based on whether or not a compound has a mutagenic effect on bacterium.

\item[PTC] \citep{toivonen:PTC} is a dataset of 344 chemical compounds (as graphs) with nineteen node features that are divided into two classes depending on whether they are carcinogenic in rats. 

\item[PROTEINS] \citep{borgwardt:PROTEINS} dataset contains 1,113 proteins (as graphs) with three node features, where the goal of the classification is to predict whether the protein is enzyme or not. 

\item[D\&D] \citep{dobson:DnD} dataset contains 1,178 protein structures (as graphs) that, similar to the previous one, are classified as enzymes or non-enzymes. 

\item [ENZYMES] \citep{borgwardt:PROTEINS} is a dataset of 600 protein structures (as graphs) with three node features. These proteins are divided into six classes of enzymes (labelled by enzyme commission numbers) for classification.

\item[COLLAB] \citep{yanardag:DGK-SOCIAL} is a scientific collaboration dataset contains 5K graphs. The classification goal here is to predict whether the graph belongs to a subfield of Physics. 

\item[IMDB-B] \citep{yanardag:DGK-SOCIAL} is a movie collaboration dataset with contains 1K graphs. The graphs are generated on two genres: Action and Romance, the classification goal is to predict the correct genre for each graph. 

\item[IMDB-M] \citep{yanardag:DGK-SOCIAL} is similar to IMDB-B, but with 1.5K graphs \& 3 genres: Comedy, Romance, and Sci-Fi. 

\item[REDDIT-B] \citep{yanardag:DGK-SOCIAL} is a dataset with 2K graphs, where each graph corresponds to an online discussion thread. The classification goal is to predict whether the graph belongs to a Q\&A-based community or discussion-based community. 

\item[REDDIT-5K] \citep{yanardag:DGK-SOCIAL} consists of 5K threads (as graphs) from five different subreddits. The classification goal is to predict the corresponding subreddit for each thread. 

\item[REDDIT-12K] \citep{yanardag:DGK-SOCIAL} is similar to REDDIT-5k, but with 11,929 graphs from 12 different subreddits. 
\end{description}
Table \ref{table: data sets} summarizes the size of available graph data (i.e., number of graphs, and both max \& mean number of vertices within graphs) in these datasets, as previously reported in the literature.
\begin{table*}[ht]
\caption{Basic statistics of the graph classification databases}
\label{table: data sets}
\begin{center}
\begin{tabular}{cccccccc} 
~ & NCI1 & NCI109 & MUTAG & D\&D & PTC & PROTEINS\\
\hline\\
\multicolumn{1}{l}{\# of graphs in data:} & 4110 & 4127 & 188 & 1178 & 344 & 1113 \\
\multicolumn{1}{l}{Max \# of vertices:} & 111 & 111 & 28 & 5748 & 109 & 620 \\
\multicolumn{1}{l}{Mean \# of vertices:} & 29.8 & 29.6 & 17.93 & 284.32 & 25.56 & 39.0 \\
\multicolumn{1}{l}{\# of features per vertex:} & 37 & 38 & 7 & 89 & 22 & 3 \\
\multicolumn{1}{l}{Mean \# of edges:} & 64.6 & 62.2 & 39.50 & 1431.3 & 51.90 & 72.82 \\
\multicolumn{1}{l}{\# of classes:} & 2 & 2 & 2 & 2 & 2 & 2 \\
\hline\hline\\
\multirow{2}{*}{ENZYMES} &\multirow{2}{*}{COLLAB} & \multicolumn{2}{c}{IMDB} & \multicolumn{3}{c}{REDDIT} \\
&  & B & M & B & 5K & 12K \\
\hline\\
600 & 5000 & 1000 & 1500 & 2000 & 5000 & 11929 \\
126 & 492 & 136 & 89 & 3783 & 3783 & 3782 \\
32.6 & 74.49 & 19.77 & 13 & 429.61 & 508.5 & 391.4 \\
3 & 3 & 3 & 3 & 2 & 2 & 2 \\
124.2 & 2457.78 & 96.53 & 65.94 & 497.75 & 594.87 & 456.89 \\
6 & 3 & 2 & 3 & 2 & 5 & 11 \\
\hline\\
\end{tabular}
\end{center}
\end{table*}

\paragraph{Graph signals for social network data:} None of the social network datasets has ready-to-use node features. Therefore, in the case of COLLAB, IMDB-B, and IMDB-M, we use the eccentricity, degree, and clustering coefficients for each vertex as characteristic graph signals. In the case of REDDIT-B, REDDIT-5K and REDDIT-12K, on the other hand, we only use degree and clustering coefficient, due to presence of disconnected graphs in these datasets.

\section{Technical Details}
\label{apx: details}

The computation of the scattering features is based on several design choices, akin to typical architecture choices in neural networks. Most importantly, it requires a choice of \begin{enumerate*} \item which statistical moments to use (normalized or unnormalized), \item the number of wavelet scales to use (given by $J$), and \item the number of moments to use (denoted by $Q$). \end{enumerate*} In general, $J$ can be automatically tuned by the diameter of the considered graphs (e.g., setting it to the logarithm of the diameter), and the other choices can be tuned via cross-validation. However, we have found the impact of such tuning to be minor, and thus for simplicity, we fix our configuration to use normalized moments, $J=5$, and $Q=4$ throughout this work.

\paragraph{Cross validation procedure:} Classification evaluation was done with standard ten-fold cross validation procedure. First, the entire dataset is randomly split into ten subsets. Then, in each iteration (or ``fold''), nine of them are used as training and validation, and the other one is used for testing classification accuracy. In total, after ten iterations, each of the subsets has been used once for testing, resulting in ten reported classification accuracy numbers for the examined dataset. Finally, the mean and standard deviation of these ten accuracies are computed and reported.

It should be noted that during training, each iteration also performs automatic tuning of the trained classifier, as follows. First, nine iterations are performed, each time using eight subsets (i.e., folds) as training and the remaining one as validation set, which is used to determine the optimal parameters for SVM. After nine iterations, each of the training/validation subsets has been used once for validation, and we obtain nine classification models, which in turn produce nine predictions (i.e., class assignments) for each data point in the test subset of the main cross validation. To obtain the final predicted class of this cross validation iteration, we select the class with the most votes (from among the nine models) as our final classification result. These results are then compared to the true labels (in the test set) on the test subset to obtain classification accuracy for this fold.

\paragraph{Software \& hardware environment:} Geometric scattering and related classification code were implemented in Python. All experiments were performed on HPC environment using an intel16-k80 cluster, with a job requesting one node with four processors and two Nvidia Tesla k80 GPUs. 

\end{appendices}

\end{document}